\documentclass[5p,twocolumn]{elsarticle}
\usepackage{hyperref, amsmath,amssymb, graphicx,subcaption}
\usepackage{color}

\journal{Neural Networks}

\makeatletter
\def\ps@pprintTitle{%
	\let\@oddhead\@empty
	\let\@evenhead\@empty
	\def\@oddfoot{}%
	\let\@evenfoot\@oddfoot}
\makeatother
\begin{document}

\begin{frontmatter}

\title{Accelerating Deep Learning with Memcomputing}

\author{Haik Manukian\fnref{myfootnote}}
\ead{hmanukia@ucsd.edu}
\author{Fabio L. Traversa\fnref{myfootnote2}} 
\ead{ftraversa@memcpu.com}
\author{Massimiliano Di Ventra\fnref{myfootnote}}
\ead{diventra@physics.ucsd.edu}
\fntext[myfootnote]{Department of Physics, University of California, San Diego, La Jolla, CA 92093}
\fntext[myfootnote2]{MemComputing, Inc., San Diego, CA, 92130 CA}







\begin{abstract}
Restricted Boltzmann machines (RBMs) and their extensions, often called ``deep-belief networks'', are powerful neural networks that have found applications in the fields of machine learning and artificial intelligence. The standard way to training these models resorts to an iterative unsupervised procedure based on Gibbs sampling, called ``contrastive divergence'', and additional supervised tuning via back-propagation. However, this procedure has been shown not to follow any gradient and can lead to suboptimal solutions. In this paper, we show 
an efficient alternative to contrastive divergence by means of simulations 
of digital memcomputing machines (DMMs) that compute the gradient of the log-likelihood involved in unsupervised training. We test our approach on pattern recognition using a modified version of the MNIST data set of hand-written numbers. DMMs sample very 
effectively the vast phase space defined by the probability distribution of RBMs over the test sample inputs, and provide a very good  
approximation close to the optimum. This efficient search significantly reduces the number of generative pretraining iterations necessary to achieve a given level of accuracy in the MNIST data set, as well as a total performance gain over the traditional approaches. In fact, the acceleration of the pretraining 
achieved by {\it simulating} DMMs is comparable to, in 
number of iterations, the recently reported {\it hardware} application of the quantum annealing method on the same network and data set. Notably, however, DMMs perform far better than the 
reported quantum annealing results in terms of {\it quality} of the training. Finally, we also compare our method to recent advances in supervised training, like batch-normalization and rectifiers, that seem to reduce the advantage of pretraining. We find that the memcomputing method still maintains a quality advantage ($>1\%$ in accuracy, corresponding to a $20\%$ reduction in error rate) 
over these approaches, despite the network pretrained with memcomputing defines a more non-convex landscape using sigmoidal activation functions without batch-normalization. Our approach is agnostic about the connectivity of the network. Therefore, it can be extended to train full Boltzmann machines, and even deep networks at once.
\end{abstract}
 
\begin{keyword}
Deep Learning \sep Restricted Boltzmann Machines\sep Memcomputing 
\end{keyword}

\end{frontmatter}


\section{Introduction}
The progress in machine learning and big data driven by successes in deep learning is difficult to overstate. Deep learning models (a subset of which are called ``deep-belief networks'') are artificial neural networks with a certain amount of layers, $n$, with $n > 2$ \cite{lecun2015deep}. They have proven themselves to be very useful in a variety of applications, from computer vision \cite{krizhevsky2012imagenet} and speech recognition \cite{hinton2012deep} to super-human performance in complex games\cite{mnih2015human}, to name just a few. While some of these models have existed for some time \cite{smolensky1986information}, the dramatic increases in computational power combined with advances in effective training methods have pushed forward these fields considerably \cite{bengio2009learning}. 

Successful training of deep-belief models relies heavily on some variant of an iterative gradient-descent procedure, called back-propagation, through the layers of the network~\cite{rumelhart1986learning}. Since this optimization method uses only gradient information, and the error landscapes of deep networks are highly non-convex \cite{choromanska2015loss}, one would at best hope to find an appropriate local minimum. 

However, there is evidence that in these high-dimensional non-convex settings, the issue is not getting stuck in some local minima but rather at saddle points, where the gradient also vanishes \cite{dauphin2014identifying}, hence making the gradient-descent procedure of limited use. A takeaway from this is that a ``good'' initialization procedure for 
assigning the weights of the network, known as {\it pretraining}, can then be highly advantageous. 

One such deep-learning framework that can utilize this pretraining procedure is the Restricted Boltzmann Machine (RBM) \cite{smolensky1986information}, and its extension, the Deep Belief Network (DBN) \cite{hinton2006fast}. These machines are a class of neural network models capable of unsupervised learning of a parametrized probability distribution over inputs. They can also be easily extended to the supervised learning case by training an output layer using back-propagation or other standard methods \cite{lecun2015deep}. 

Training RBMs usually distinguishes between an unsupervised pretraining, whose purpose is to initialize a good set of weights, and the supervised procedure. The current most effective technique for pretraining RBMs utilizes an iterative sampling technique called {\it contrastive divergence} (CD) \cite{hinton2006training}. Computing the exact gradient of the log-likelihood is exponentially hard in the size of the RBM, and so CD approximates it with a computationally friendly sampling procedure. While this procedure has brought RBMs most of their success, CD suffers from the slow mixing of Gibbs sampling, and is known not to follow the gradient of any function \cite{sutskever2010convergence}.

Partly due to these shortcomings of pretraining with CD, much research has gone into making the back-propagation procedure more robust and less sensitive to the initialization of weights and biases in the network. This includes research into different non-linear activation functions (e.g., ``rectifiers'') \cite{glorot2011deep} to combat the vanishing gradient problem and normalization techniques (such as ``batch-normalization'') \cite{pmlr-v37-ioffe15} that make back-propagation in deep networks more stable and less dependent on initial conditions. In sum, these techniques make training deep networks an easier (e.g., more convex) optimization problem for a gradient-based approach like back-propagation. This, in turn, relegates the standard CD pretraining procedure's usefulness to cases where the training set is sparse \cite{lecun2015deep}, which is becoming an increasingly rare occurrence.

In parallel with this research into back-propagation, sizable effort has been expended toward improving the power of the pretraining procedure, including extensions of CD \cite{tieleman2009using, 2018arXiv180102567R}, CD done on memristive hardware \cite{sheri2015contrastive}, and more recently, approaches based on quantum annealing that try to recover the exact gradient \cite{biamonte2017quantum} involved in pretraining. Some of these methods are classical algorithms simulating quantum sampling \cite{wiebe2014quantum}, and still others attempt to use a {\it hardware} quantum device in contact with an environment to take independent samples from its Boltzmann distribution for a more accurate gradient computation. For instance, in a recent work, the state of the RBM has been mapped onto a commercial quantum annealing processor (a D-Wave machine), the latter used as a sampler of the model distribution~\cite{adachi2015application}. The results reported on a reduced version of the well-known MNIST data set look promising as compared to CD~\cite{adachi2015application}. However, these approaches require expensive hardware, and cannot be scaled to larger problems as of yet. 

In the present paper, inspired by the theoretical underpinnings~\cite{umm,dmm} and recent empirical demonstrations \cite{exponential2017speedup, manukian2017inversion} of the advantages of a new computing paradigm --{\it memcomputing}~\cite{thepa}-- on a variety of combinatorial/optimization problems, we seek to test its power toward the computationally demanding problems in deep learning. 

Memcomputing~\cite{thepa} is a novel computing paradigm that solves complex computational problems using processing embedded in memory. It has been formalized by two of us (FLT and MD) by introducing the concept of universal memcomputing machines\cite{umm}. In short, to perform a computation, the task at hand is mapped to a continuous dynamical system that employs highly-correlated states \cite{topo} (in both space and time) of the machine to navigate the phase space efficiently and find the solution 
of a given problem as mapped into the equilibrium states of the dynamical system. 

In this paper, we employ a subset of these machines called {\it digital memcomputing machines} (DMMs) and, more specifically, their self-organizing circuit realizations \cite{dmm,Di_Ventra2018}. The distinctive feature of DMMs is their ability to read and write the initial and final states of the machine {\it digitally}, namely 
requiring only {\it finite} precision. This feature makes them easily {\it scalable} as our modern computers. 

From a practical point of view DMMs can be built with standard circuit elements with and without memory~\cite{dmm}. These elements, however, are {\it non-quantum}. 
Therefore, the ordinary differential equations of the corresponding circuits can be efficiently simulated on our present computers. Here, we will indeed employ 
only {\it simulations} of DMMs on a single Xeon processor to train RBMs. These simulations show already substantial advantages with respect to CD and even quantum annealing, despite the latter is 
executed on hardware. Of course, the hardware implementation of DMMs applied to these problems would offer even more advantages since the simulation times 
will be replaced by the actual physical time of the circuits to reach equilibrium. This would then offer a realistic path to real-time pretraining of deep-belief networks. 

In order to compare directly with quantum annealing results recently reported\cite{adachi2015application}, we demonstrate the advantage of our memcomputing approach by first training on a reduced MNIST data set as that used in Ref~\cite{adachi2015application}. We show that our method requires far less pretraining iterations to achieve the same accuracy as CD, as well as an overall accuracy gain over both CD and quantum annealing. We also train the RBMs on the reduced MNIST data set 
without mini-batching, where the quantum annealing results are not available. Also in this case, we find both a substantial reduction in pretraining iterations needed as well as a higher level of accuracy of the 
memcomputing approach over the traditional CD. 

Our approach then seems to offer many of the advantages 
of quantum approaches. However, since it is based on a completely classical system, it can be efficiently deployed in software (as we demonstrate in this paper) as 
well as easily implemented in hardware, and can be scaled to full-size problems. 

Finally, we investigate the role of recent advances in supervised training by comparing accuracy obtained using only back-propagation with batch-normalization and rectifiers 
starting from a random initial condition versus the back-propagation procedure initiated from a network pretrained with memcomputing, {\it with} sigmoidal activations and {\it without} batch-normalization. Even without these advantages, namely operating on a more non-convex landscape, we find the network pretrained with memcomputing maintains an accuracy gain over state-of-the-art back-propagation by more than 1$\%$ and a $20\%$ reduction in error rate. This gives further evidence to the fact that memcomputing pretraining navigates to an advantageous initial point in the non-convex loss surface of the deep network.

\section{RBMs and Contrastive Divergence}

\begin{figure}[t]
	\centering
	\includegraphics[width=8.5cm
	]{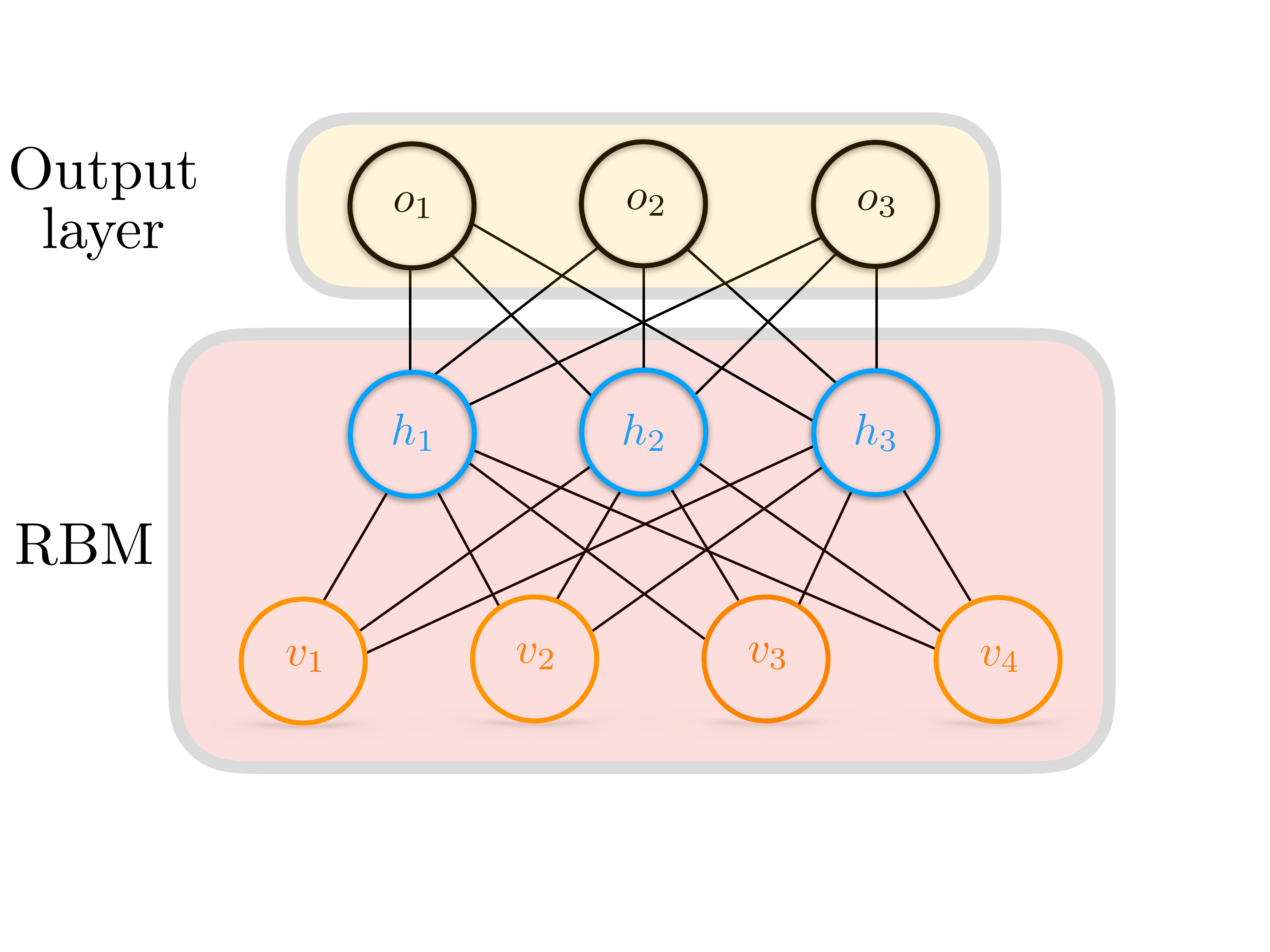}
	\caption{A sketch of an RBM with four visible nodes, three hidden nodes, and an output layer with three nodes. The value of each stochastic binary node is represented by $v_i, h_i \in \{0,1\}$, which are sampled from the probabilities in Eqs.~(\ref{hs}),~(\ref{vs}). The connections between the layers represent the weights, $w_{ij} \in \mathbb{R}$ (biases not shown). Note the lack of connections between nodes in the same layer, which distinguishes the RBM from a Boltzmann machine. The RBM weights are trained separately from the output layer with generative pretraining, then tuned together via back-propogation (just as in a feed-forward neural network).}
	\label{figmodel}
\end{figure}

An RBM consists of $m$ visible units, $v_j, j=1\dots m$, each fully connected to a layer of $n$ hidden units, $h_i, i=1\dots n$, both usually taken to be binary variables. In the restricted model, no intra-layer connections are allowed, see Fig.~\ref{figmodel}. 

The connectivity structure of the RBM implies that, given the hidden variables, each input node is conditionally independent of all the others:
\begin{equation}
p(v_i, v_j | {\bf h}) = p(v_i |{\bf h}) p(v_j | {\bf h}).
\end{equation}
The joint probability is given by the Gibbs distribution,
\begin{equation}
p({\bf v}, {\bf h}) = \frac{1}{Z} e^{-E({\bf v},{\bf h})},
\end{equation}
with an energy function 
\begin{equation}
\label{energy}
E({\bf v}, {\bf h}) = - \sum_i \sum_j w_{ij}h_iv_j -\sum_j b_jv_j - \sum_i c_i h_i,
\end{equation}
where $w_{ij}$ is the weight between the $i$-th hidden neuron and the $j$-th visible neuron, and $b_j$, $c_i$ are real numbers indicating the ``biases'' 
of the neurons. The value, $Z$, is a normalization constant, and is known in statistical mechanics as the partition function.
Training an RBM then amounts to finding a set of weights and biases that maximizes the likelihood (or equivalently minimizes the energy) of the observed data.   

A common approach to training RBMs for a supervised task is to first perform generative unsupervised learning (pretraining) to initialize the weights and biases, then run back-propagation over input-label pairs to fine tune the parameters of the network. The pretraining is framed as a gradient ascent over the log-likelihood of the observed data, which gives a particularly tidy form for the weight updates from the $n$-th to the $(n+1)$-th iteration:
\begin{equation}
\label{gradient}
\Delta w^{n+1}_{ij} = \alpha \Delta w^n_{ij}+\epsilon [\langle v_ih_j \rangle_{\text{DATA}} - \langle v_i h_j \rangle_{\text{MODEL}}],
\end{equation}  
where $\alpha$ is called the ``momentum'' and $\epsilon$ is the ``learning rate''. A similar update procedure is applied to the biases:
\begin{equation}
\label{gradientb}
\Delta b^{n+1}_{i} = \alpha \Delta b^n_{i}+\epsilon [\langle v_i \rangle_{\text{DATA}} - \langle v_i \rangle_{\text{MODEL}}],
\end{equation}
\begin{equation}
\label{gradientc}
\Delta c^{n+1}_{j} = \alpha \Delta c^n_{j}+\epsilon [\langle h_j \rangle_{\text{DATA}} - \langle h_j \rangle_{\text{MODEL}}].
\end{equation}

This form of the weight updates is referred to as ``stochastic gradient optimization with momentum''. The first expectation value on the rhs of Eqs.~(\ref{gradient}),~(\ref{gradientb}), and~(\ref{gradientc}) is taken with respect to the conditional probability distribution with the data fixed at the visible layer. This is relatively easy to compute. Evaluation of the second expectation on the rhs of Eqs.~(\ref{gradient}),~(\ref{gradientb}), and~(\ref{gradientc}) is exponentially hard in the size of the 
network, since obtaining independent samples from a high-dimensional model distribution easily becomes prohibitive with increasing size \cite{hinton2006training}. This is the term that CD attempts to approximate. 

The CD approach attempts to reconstruct the difficult expectation term with iterative Gibbs sampling. This works by sequentially sampling each layer given the sigmoidal conditional probabilities, namely
\begin{equation}
\label{hs}
p(h_i = 1 | {\bf v}) = \sigma \left(\sum_{j}w_{ij}v_j + c_i\right), 
\end{equation}
for the visible layer, and similarly for the hidden layer
 \begin{equation}
\label{vs}
p(v_j = 1 | {\bf h}) = \sigma \left(\sum_{i}w_{ij}h_i + b_j\right), 
\end{equation}
with $\sigma(x) = (1 + e^{-x})^{-1}$. The required expectation values are calculated with the resulting samples. In the limit of infinite sampling iterations, the expectation value is recovered. However, this convergence is slow and in practice usually only one iteration, 
referred to CD-1, is used \cite{hinton2010practical}.

\section{Efficient Sampling with Memcomputing}\label{efficient}
\subsection{The memcomputing approach to optimization}
In this work, we propose the application of DMMs to the accurate training of restricted Boltzmann machines. Formally, a DMM can be specified by the following tuple,~\cite{dmm}
\begin{equation}
\text{DMM} = (\mathbb{Z}_2, \Delta, \mathcal{P}, S, \Sigma, p_0, s_0, F)
\end{equation}
Where $\Delta$ is the set of transition functions between states of the machine,

\begin{equation}
\delta_{\alpha} : \mathbb{Z}^{m_{\alpha}}_2 \setminus F \times \mathcal{P} \to \mathbb{Z}_2^{m_{\alpha}'}\times \mathcal{P}^2 \times S
\end{equation}
Here $m_\alpha$ represents the number of memprocessors read in by $\delta_\alpha$ and $m_\alpha'$ the number of memprocessors written by the function. $\mathcal{P}$ is the set of arrays of pointers $p_\alpha$ that identify memprocessors called by $\delta_\alpha$, $S$ is the collection of indices $\alpha$, $\Sigma$ is a set of initial states, $p_0$ an initial array of pointers, $s_0$ an initial index and $F$ a set of final or accepting states.

A practical realization of DMMs can be accomplished using dynamical systems~\cite{dmm,Di_Ventra2018}. In this paper, we employ the representation that uses self-organizing logic circuits (SOLCs), namely circuits that self-organize to satisfy the appropriate logical propositions defined by the 
	problem at hand. These circuits are fully specified by a set of coupled ordinary differential equations representing the physical system of electric components comprising the circuit,

\begin{equation}
\label{solc_eqn}
\dot{y} = G(y(t)),
\end{equation}
where $y$ describes the vector of all voltages, currents and internal state variables (providing memory) in the circuit, and $G$ is the flow vector field 
that defines the laws of temporal evolution of the circuit.

To solve a given computational problem within this paradigm, first a Boolean circuit is constructed that represents the given problem. The constituent classical logic gates are then replaced with ones that self-organize~\cite{dmm}. The inputs to the problem can be specified via voltages at terminals of the SOLC, whose equations of motion are then integrated forward in time from a random initial condition, according to the vector flow field $F$, to an equilibrium (fixed) point of the system. Voltage terminals corresponding to the solution of the problem can then be read out.

\emph{Prima facie}, the mapping of a discrete Boolean problem into a continuous system of non-linear differential equations may not seem computationally advantageous. In this case, however, the mapping to a SOLC results in a vector flow field which has been shown to possess certain functional and topological properties that give rise to dramatic computational advantages~\cite{dmm,Di_Ventra2018}. Namely, the resulting dynamical systems do not exhibit chaos or periodic orbits in the presence of solutions~\cite{no-chaos,noperiod}, and employ highly non-local (in both space and time) correlated states to efficiently explore the vast phase space of the problem~\cite{topo,Bearden2018}. It is the latter property that truly distinguishes this approach from other, local, approaches to optimization.

\begin{figure}[t]
	\centering
	\includegraphics[width=8.05cm
	]{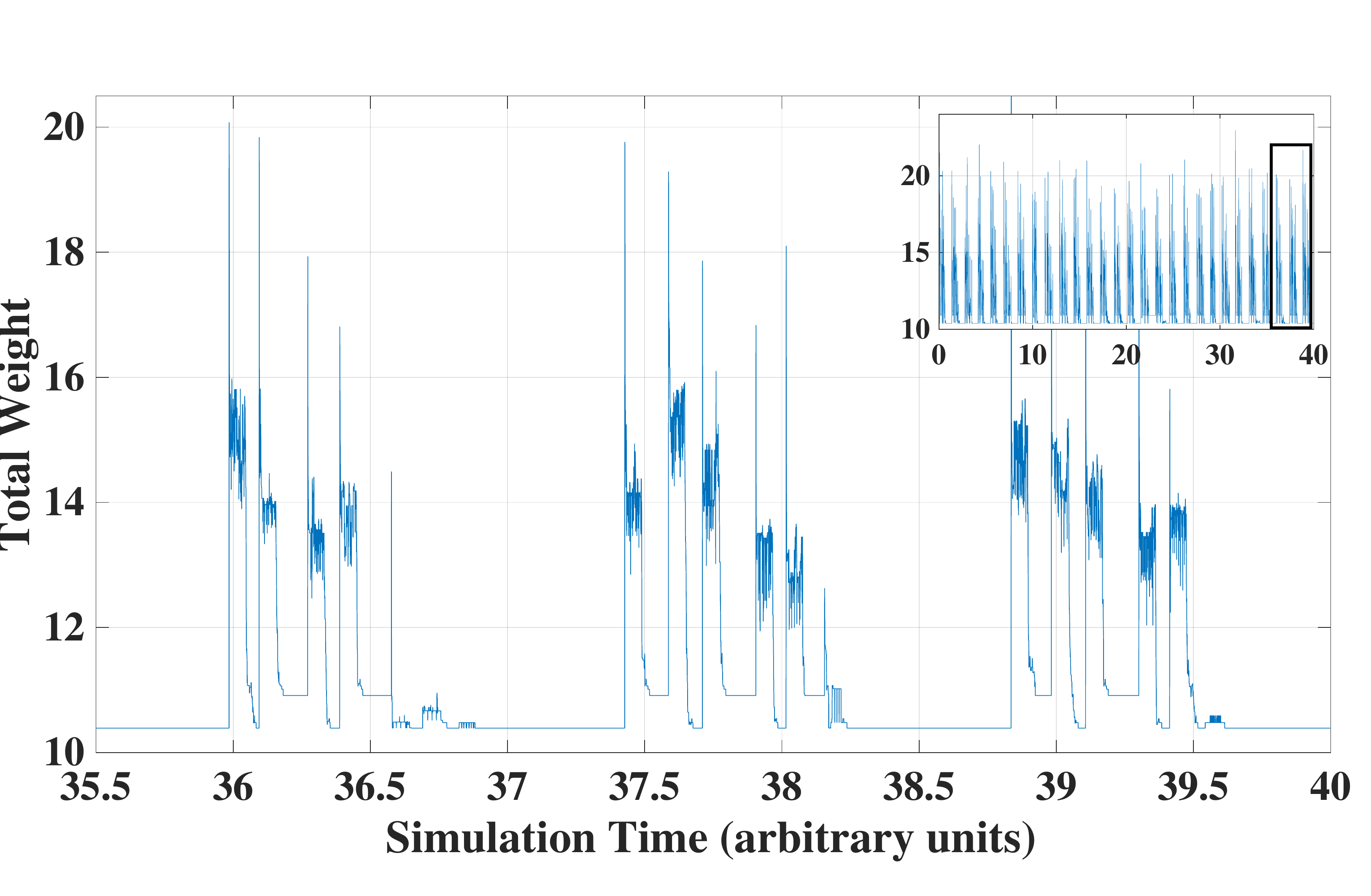}
	\caption{Plot of the total weight of a MAX-SAT clause as a function of internal simulation time (not physical seconds) of a DMM. A lower weight variable assignment corresponds directly to a higher probability assignment of the nodes of an RBM. If the simulation has not changed assignments in some time, we restart with another random (independent) initial condition. The inset shows the full simulation, with all restarts. The main figure focuses on the last three restarts, signified by the black box in the inset.}
	\label{fig:trajectory}
\end{figure}

\begin{figure}[t]
	\centering
	\includegraphics[width=9.55cm
	]{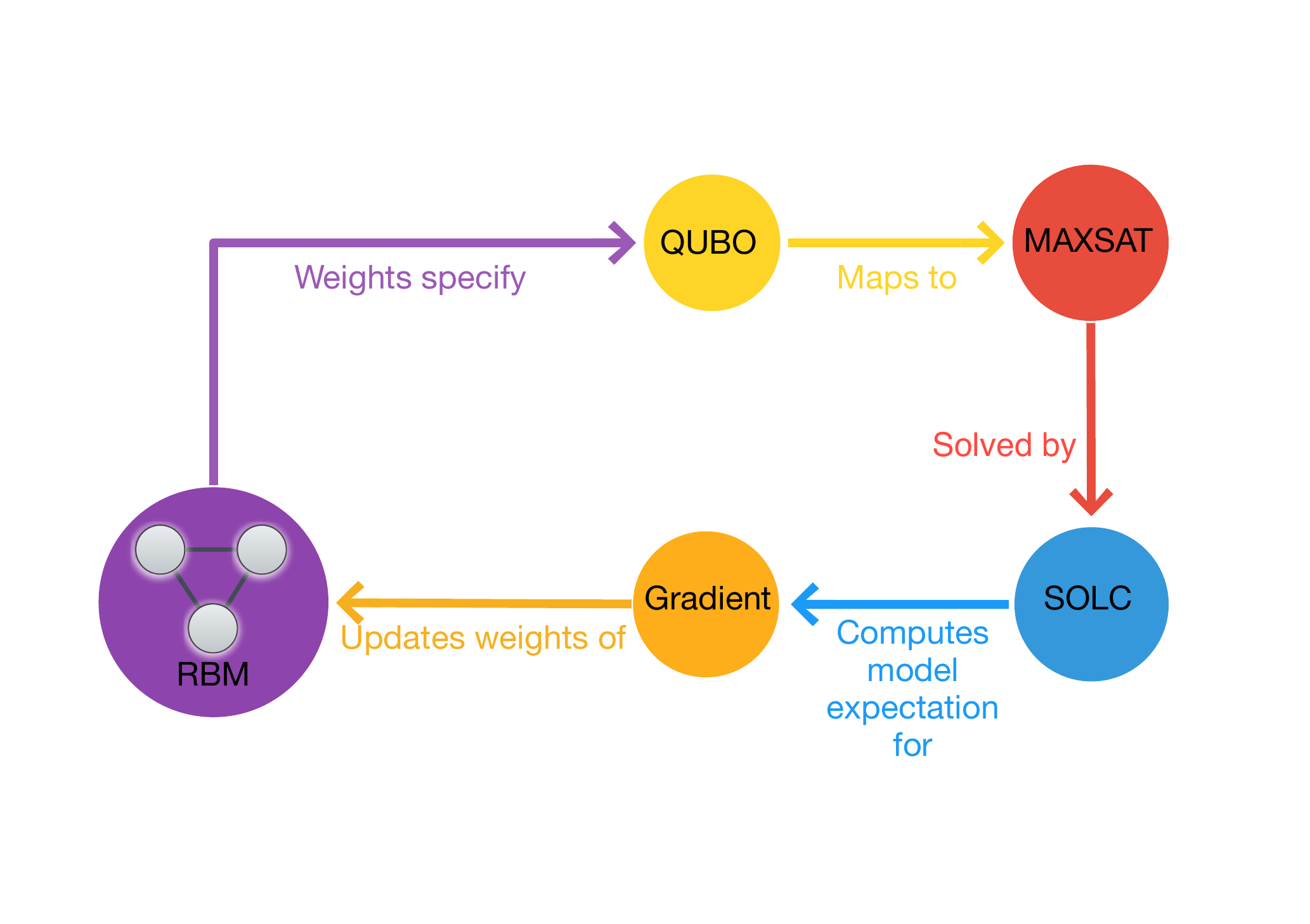}
	\caption{A diagrammatic overview of the memcomputing-assisted RBM training procedure described in this paper. For each iteration of pretraining, the set of weights of an RBM specifies a QUBO problem, which is converted into an equivalent weighted MAX-SAT problem to be solved by a DMM in its SOLC representation. This solution computes the model expectation term in the gradient~(\ref{gradient}), which updates the weights of the model and the whole process begins again, until convergence.}
	\label{fig:flowchart}
\end{figure}
\subsection{From RBMs to a QUBO problem to a MAX-SAT}

To employ these advantages within the training of RBMs, we construct first a reinterpretation of the RBM pretraining that explicitly shows how it corresponds to an NP-hard 
optimization problem, which we then tackle using DMMs in their SOLC representation. We first observe that to obtain a sample near most of the probability mass of the joint distribution, $p(v,h) \propto e^{-E(v,h)}$, one must find the minimum of the energy of the form Eq.~(\ref{energy}), which constitutes a {\it quadratic unconstrained binary optimization} (QUBO) problem~\cite{computational_complexity_book}. 

We can see this directly by considering the visible and hidden nodes as one vector ${\bf x} = ({\bf v}, {\bf h})$ and re-writing the energy of an RBM configuration as 
\begin{equation}
E = -x^T Q x, 
\end{equation} 
where $Q$ is the matrix 
\begin{equation}
Q = \begin{bmatrix}
    B & W \\
    0 & C 
\end{bmatrix},
\end{equation}
with $B$ and $C$ being the diagonal matrices representing the biases $b_j$ and $c_i$, respectively, while the matrix $W$ contains the weights.

\begin{figure}[h!]
	\centering
	\label{fig:cdvsmem}
	
	\begin{subfigure}[b]{0.89\columnwidth}
		\caption{100 back-propagation iterations}
		\includegraphics[width=\columnwidth]{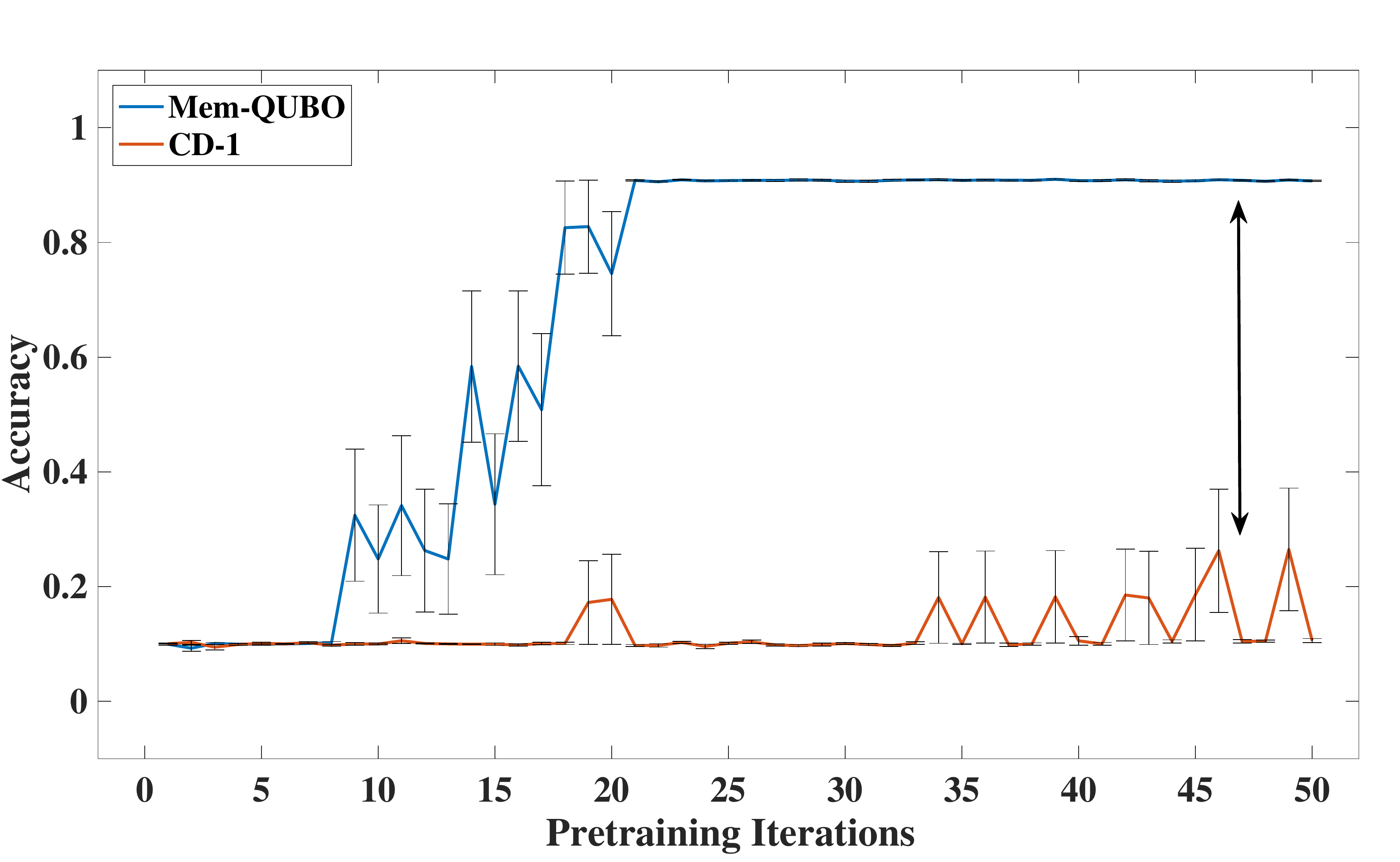}
		\label{fig:bp100}
	\end{subfigure}
	
	\begin{subfigure}[b]{0.89\columnwidth}
		\caption{200 back-propagation iterations}
		\includegraphics[width=\columnwidth]{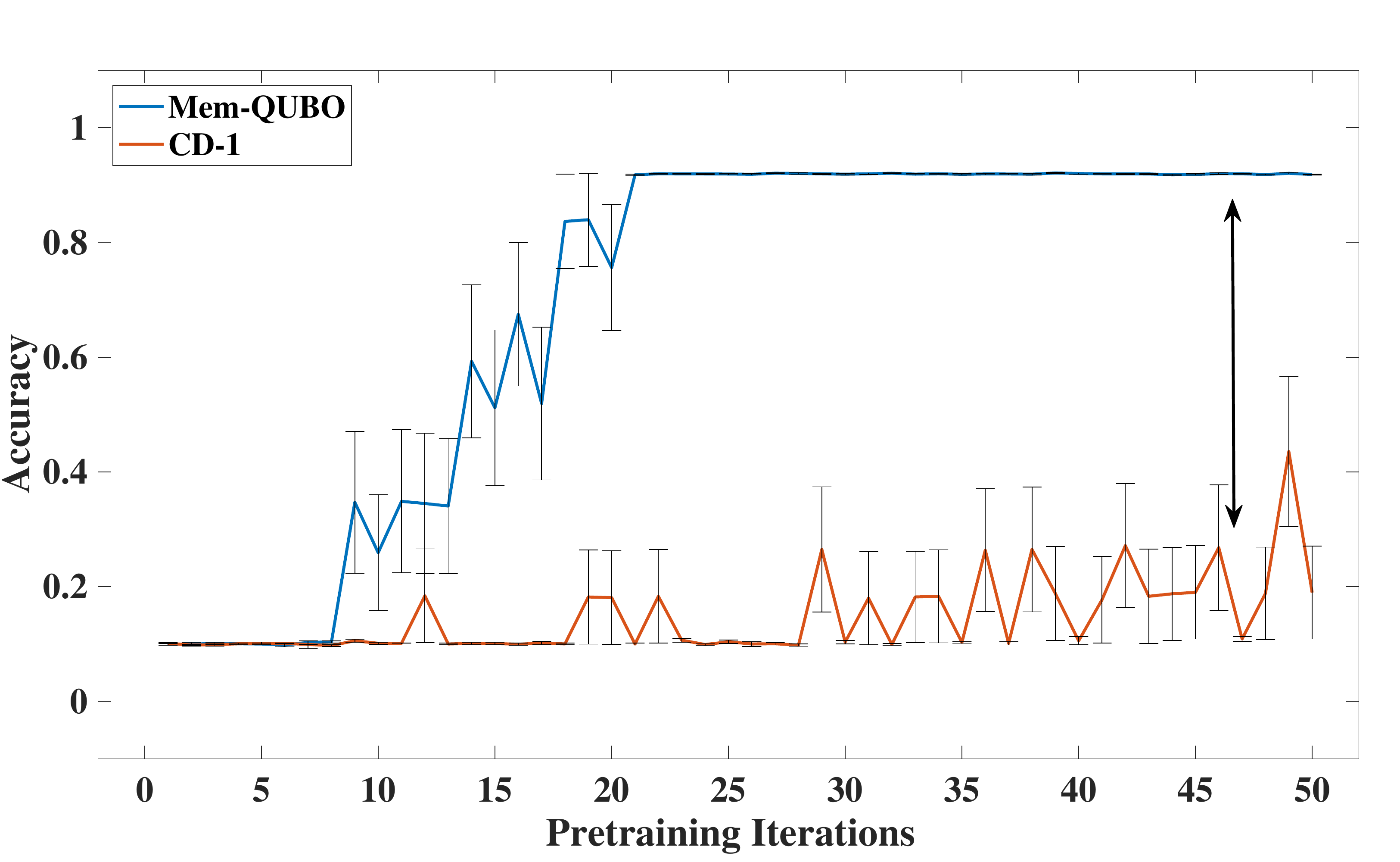}
		\label{fig:bp200}
	\end{subfigure}
	
	\begin{subfigure}[b]{0.89\columnwidth}
		\caption{400 back-propagation iterations}
		\includegraphics[width=\columnwidth]{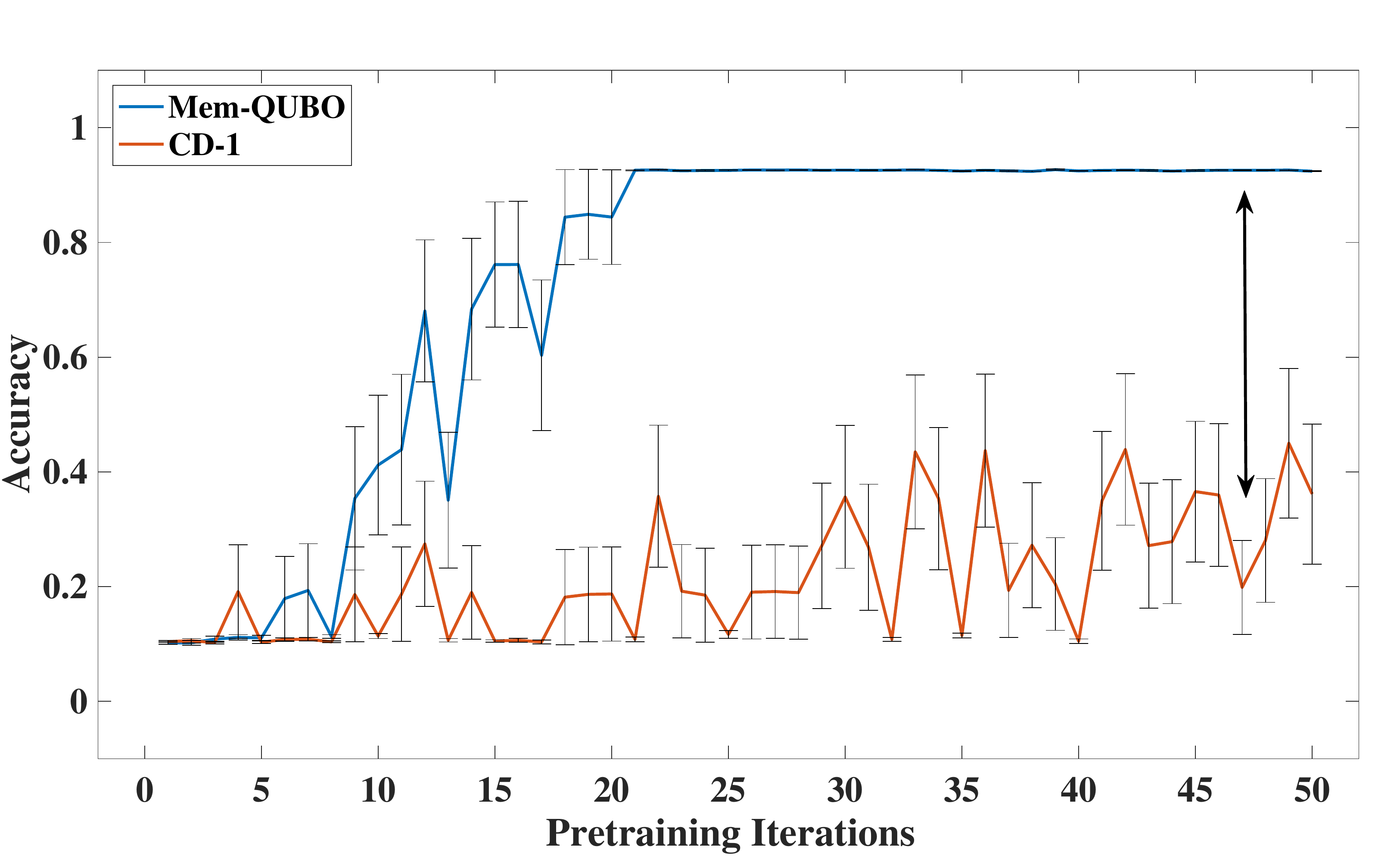}
		\label{fig:bp400} 
	\end{subfigure}
	
	\caption[Numerical Solutions]{Memcomputing (Mem-QUBO) accuracy on the test set of the reduced MNIST problem versus contrastive divergence for $n=100 \textrm{(a)}, 200 \textrm{(b)}, 400 \textrm{(c)}$ iterations of back-propagation with mini-batches of 100. The plots show average accuracy with $\pm \sigma/\sqrt{N}$ error bars calculated across 10 DBNs trained on $N=10$ different partitions of the training set. One can see a dramatic acceleration with the memcomputing approach needing far less iterations to achieve the same accuracy, as well as an overall performance gap (indicated by a black arrow) that back-propagation cannot seem to overcome. Note that some of the error bars for 
		both Mem-QUBO and CD-1 are very small on the reported scale for a number of pretraining iterations larger than about 20.}
	\label{fig:cdvsmem}
\end{figure}

We then employ a mapping from a general QUBO problem to a {\it weighted maximum satisfiability} (weighted MAX-SAT) problem, similar to \cite{bian2010ising}, which is directly solved by the DMM. The weighted MAX-SAT problem is to find an assignment of boolean variables that minimizes the total weight of a given boolean expression written in conjunctive normal form~\cite{computational_complexity_book}. 

This problem is a well-known problem in the NP-hard complexity class~\cite{computational_complexity_book}. However, it was recently shown in Ref.~\cite{exponential2017speedup}, that simulations of DMMs show dramatic (exponential) speed-up over the state-of-the-art solvers, when attempting to find better approximations to hard MAX-SAT instances beyond the inapproximability gap \cite{haastad2001some}.  We then use SOLCs appropriately designed to 
tackle the MAX-SAT that originates from the RBM QUBO problem. The ordinary differential equations we solve can be found in Ref.~\cite{dmm} appropriately 
adapted to deal with the particular problem discussed in this paper.  

The approximation to the global optimum of the 
weighted MAX-SAT problem given by memcomputing is then mapped back to the original variables that represent the states of the RBM nodes. Finally, we obtain an approximation to the ``ground state'' (lowest energy state) of the RBM as a variable assignment, ${\bf x}^*$, close to the peak of the probability distribution, where $\nabla P({\bf x}^*)=0$. This assignment is obtained by integration of the ordinary differential equations that define the SOLC's dynamics. In doing so we collect an entire trajectory, ${\bf x}(t)$, that begins at a random initial condition in the phase space of the problem, and ends at the lowest energy configuration of the variables (see Fig.~\ref{fig:trajectory}). 
\begin{figure}[h!]
	\centering
	\label{fig:cdvsmem2}
	
	\begin{subfigure}[b]{0.89\columnwidth}
		\caption{100 back-propagation iterations}
		\includegraphics[width=\columnwidth]{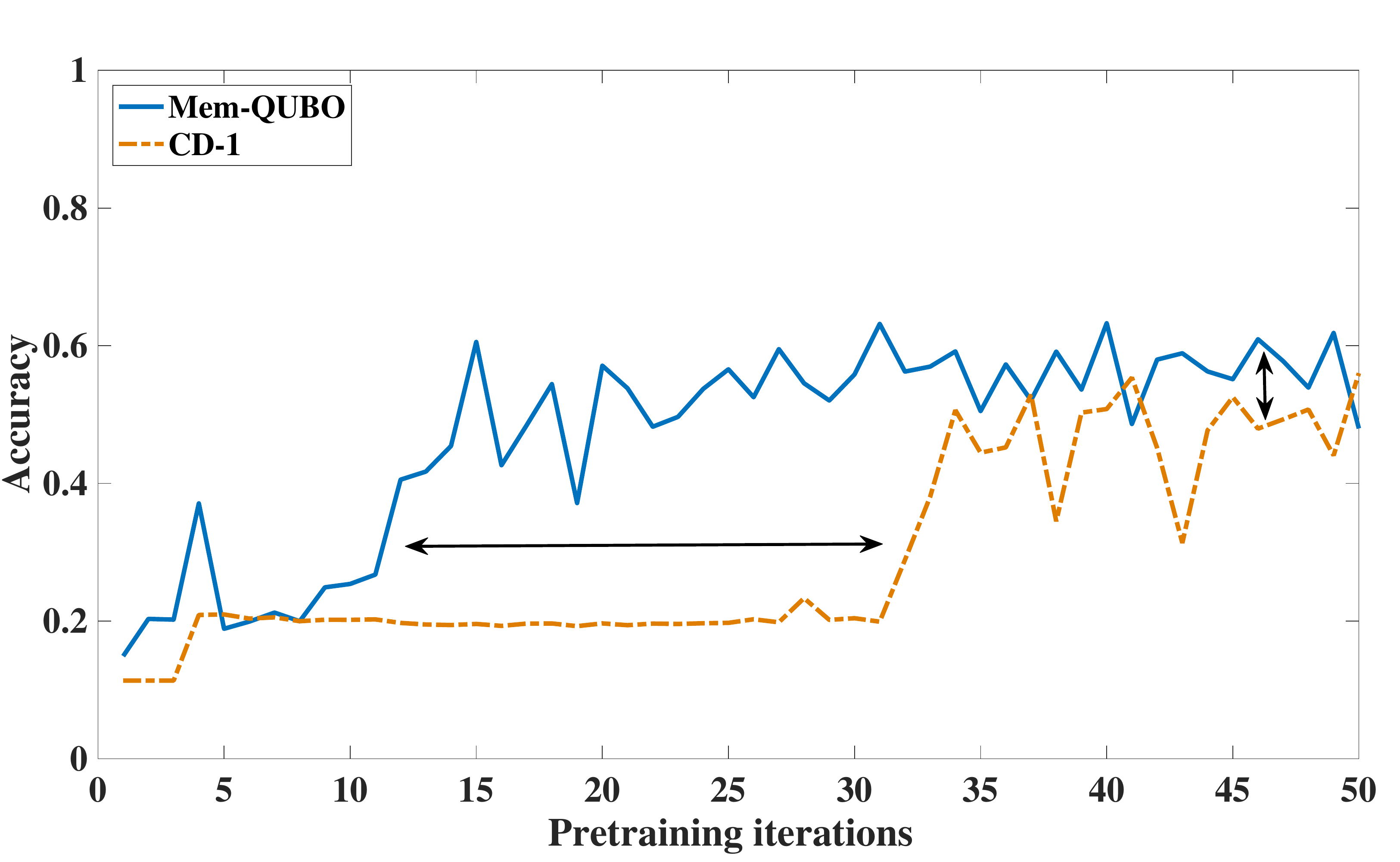}
		\label{fig:bp100-2}
	\end{subfigure}
	
	\begin{subfigure}[b]{0.89\columnwidth}
		\caption{500 back-propagation iterations}
		\includegraphics[width=\columnwidth]{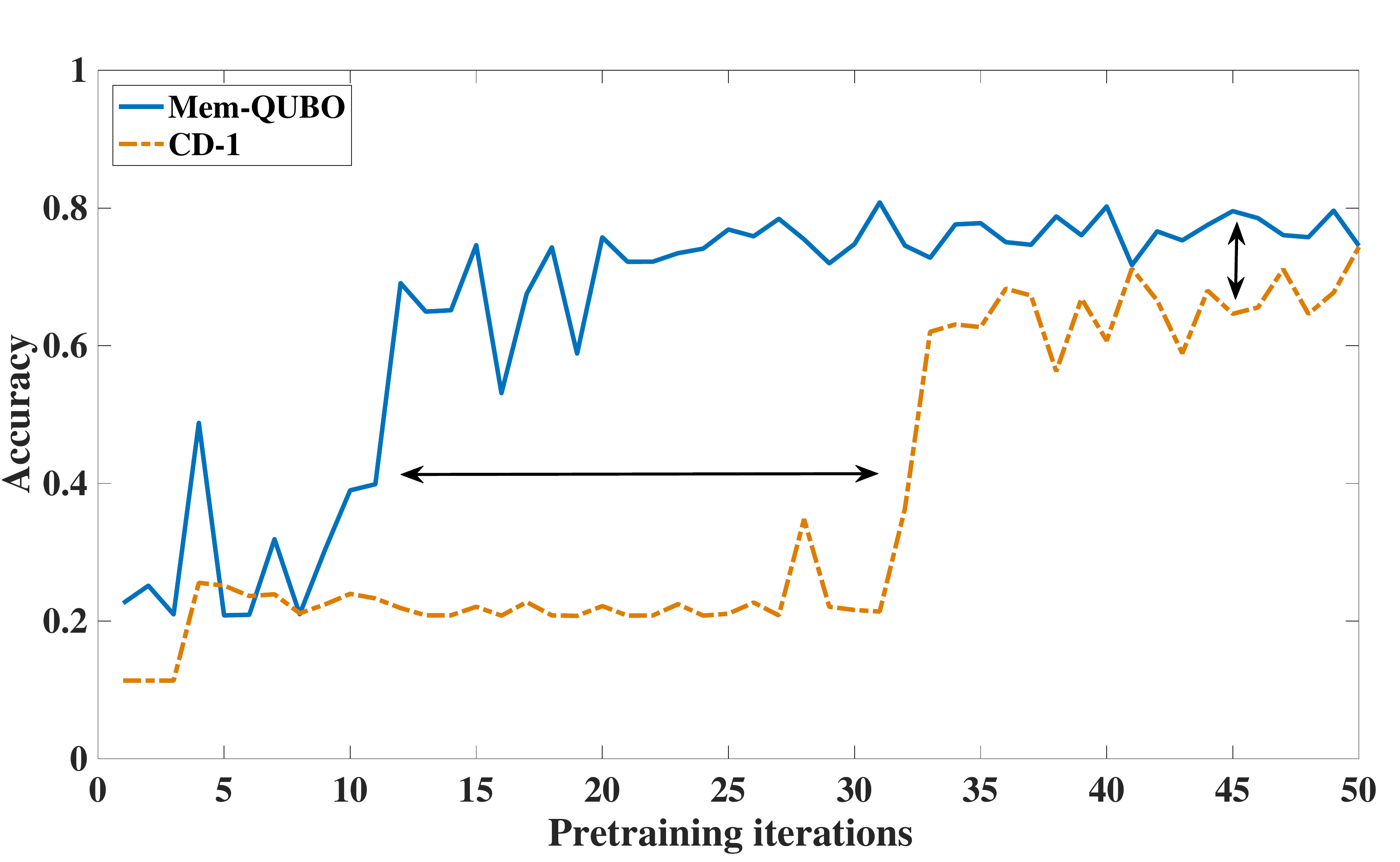}
		\label{fig:bp200-2}
	\end{subfigure}
	
	\begin{subfigure}[b]{0.89\columnwidth}
		\caption{800 back-propagation iterations}
		\includegraphics[width=\columnwidth]{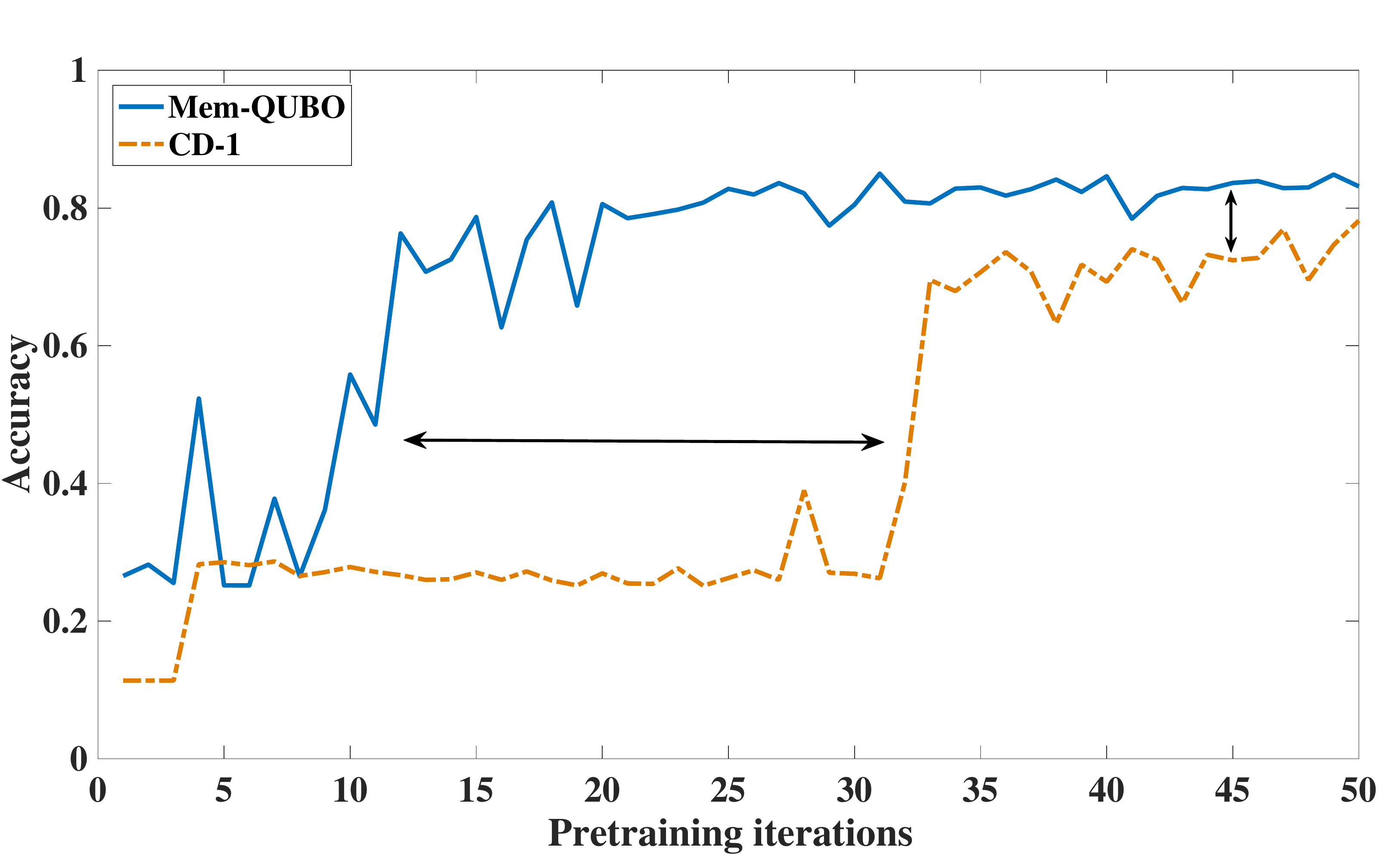}
		\label{fig:bp400-2} 
	\end{subfigure}
	
	\caption{Mem-QUBO accuracy on the reduced MNIST test set vs. CD-1 after $n=100 \textrm{(a)}, 500 \textrm{(b)}, 800 \textrm{(c)}$ iterations of back-propagation with no mini-batching. The resulting pretraining acceleration shown by the memcomputing approach is denoted by the horizontal arrow. A performance gap also appears, emphasized by the vertical arrow, with Mem-QUBO obtaining a higher level of accuracy than CD-1, even for the highest number of back-propagation iterations. No error bars appear here since we have trained the full test set.}
	\label{fig:cdvsmem2}
\end{figure}

Since the problem we are tackling here is an optimization one, we do not have any guarantee of finding the {\it global} optimum. (This is in contract to a SAT problem where 
we can guarantee DMMs {\it do} find the solutions of the problem corresponding to equilibrium points, if these exist~\cite{dmm,noperiod,no-chaos}.) Therefore, there is an ambiguity about what exactly constitutes the stopping time of the simulation, since {\it a priori}, one cannot know that the simulation has reached the {\it global} minimum. 

We then perform a few ``restarts'' of the simulation (that effectively correspond to a change of the initial conditions) and stop the simulation when the machine has not found any better configuration within that number of restarts. The restarts are clearly seen in Fig. \ref{fig:trajectory} as spikes in 
the total weight of the boolean expression. In this work we have employed 28 restarts, which is an over-kill since a much smaller number would have given similar results. 

The full trajectory, ${\bf x}(t)$, together with the above ``restarts'' is plotted in Fig. \ref{fig:trajectory}. It is seen that this trajectory, in between restarts, spends most of its time in ``low-energy regions,'' or equivalently areas of high probability. A time average, $\langle {\bf x}(t) \rangle$, gives a good approximation to the required expectations in the gradient calculation in Eqs.~(\ref{gradient}),~(\ref{gradientb}), and~(\ref{gradientc}). In practice, even using the best assignment found, ${\bf x}^*$, shows a great improvement over CD in our experience. This is what we report in this paper. Note also that a full trajectory, as the one shown in Fig.~\ref{fig:trajectory}, takes about $0.5$
seconds on a single Xeon processor. A schematic of the full pre-training iteration procedure used in this paper can be seen in the diagram in Fig.~\ref{fig:flowchart}.

\section{Results}

As a testbed for the memcomputing advantage in deep learning, and as a direct comparison to the quantum annealing hardware approaches, we first looked to the 
reduced MNIST data set as reported in \cite{adachi2015application} for quantum annealing using a D-wave machine. Therefore, we have first applied the same reduction to the full MNIST problem as given in that work, which consists of removing two pixels around all $28\times 28$ grayscale values in both the test and training sets. Then each $4\times 4$ block of pixels is replaced by their average values to give a $6\times 6$ reduced image. Finally, the four corner pixels are discarded resulting in a total of 32 pixels representing each image.

We also trained the same-size DBN consisting of two stacked RBMs each with 32 visible and hidden nodes, training each RBM one at a time. We put both the CD-1 and our Memcomputing-QUBO (Mem-Qubo) approach through $N=1, \cdots, 50$ generative pretraining iterations using no mini-batching. 

For the memcomputing approach, we solve one QUBO problem per pretraining iteration to compute the model expectation value in Eqs.~(\ref{gradient}),~(\ref{gradientb}), and~(\ref{gradientc}). We pick out the best variable assignment, ${\bf x}^*$, which gives the ground state of Eq.~(\ref{energy}) as an effective approximation of the required expectation. After generative training, an output classification layer with 10 nodes was added to the network (see Fig.~\ref{figmodel}) and 1000 back-propagation iterations were applied in both approaches using mini-batches of 100 samples to generate Fig.~\ref{fig:cdvsmem}. For both pretraining and back-propagation, our learning rate was set to $\epsilon = 0.1$ and momentum parameters were $\alpha = 0.1$ for the first 5 iterations, and $\alpha = 0.5$ for the rest, same as in \cite{adachi2015application}.

Accuracy on the test set versus CD-1 as a function of the number of pretraining iterations is seen in Fig.~\ref{fig:cdvsmem}. The memcomputing method reaches a far better solution faster, and maintains an advantage over CD even after hundreds of back-propagation iterations. Interestingly, our {\it software} approach is even competitive with the quantum annealing method done in {\it hardware} \cite{adachi2015application} (cf. Fig.~\ref{fig:cdvsmem} with Figs. 7, 8, and 9 in Ref.~\cite{adachi2015application}). This is quite a remarkable result, since we integrate a set of differential equations of a {\it classical} system, in a scalable way, with comparable sampling power to a physically-realized system that takes advantage of {\it quantum} effects to improve on CD. 

Finally, we also trained the RBM on the reduced MNIST data set without mini-batches. We are not aware of quantum-annealing results for the full data set, but we can still compare with 
the CD approach. We follow a similar procedure as discussed above. In this case, however, no mini-batching was used for a more direct comparison between the Gibbs sampling of CD and our memcomputing approach. The results are shown in Fig.~\ref{fig:cdvsmem2} for different numbers of back-propagation iterations. Even on the full modified MNIST 
set, our memcomputing approach requires a substantially lower number of pretraining iterations to achieve a high accuracy and, additionally,  
shows a higher level of accuracy over the traditional CD, even after 800 back-propagation iterations. 

\section{The Role of Supervised Training}
The computational difficulty of computing the exact gradient update in pretraining, combined with the inaccuracies of CD, has inspired research into methods which reduce (or outright eliminate) the role of pretraining deep models. These techniques include changes to the numerical gradient procedure itself, like adaptive gradient estimation \cite{kingma2014adam}, changes to the activation functions (e.g., the introduction of rectifiers) to reduce gradient decay and enforce sparsity \cite{glorot2011deep}, and techniques like batch normalization to make back-propagation less sensitive to initial conditions \cite{pmlr-v37-ioffe15}. With these new updates, in many contexts, deep networks initialized from a random initial condition are found to compete with networks pretrained with CD \cite{glorot2011deep}.

To complete our analysis we have then compared a network pretrained with our memcomputing approach to these back-propagation methods with no pretraining. Both networks were trained with stochastic gradient descent with momentum and the same learning rates and momentum parameter we used in Section~\ref{efficient}.

\begin{figure}[t]
	\centering
	\includegraphics[width=8.5cm
	]{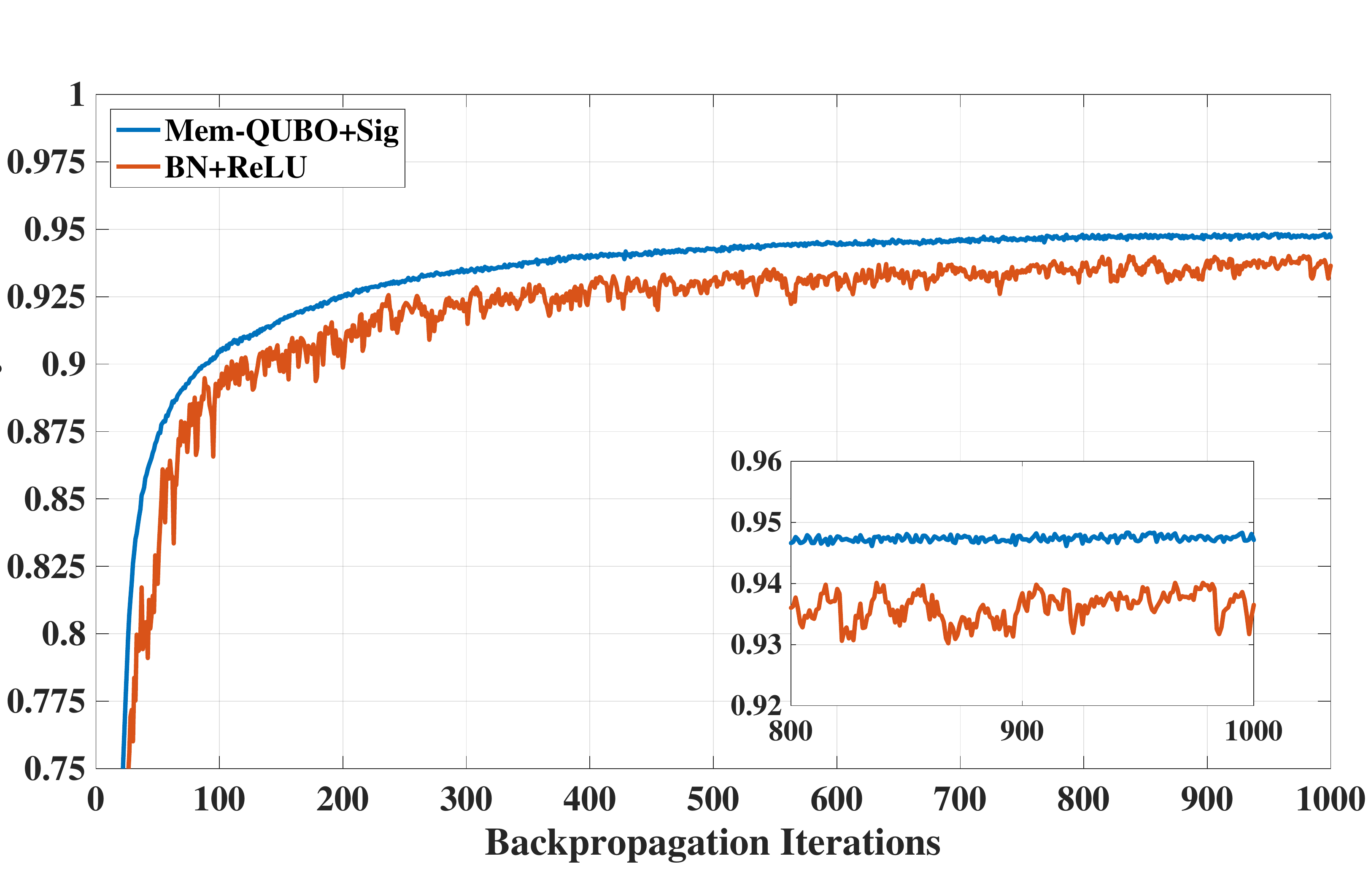}
	\caption{Accuracy on the reduced MNIST test set obtained on a network pretrained with (blue curve) our approach Mem-QUBO and sigmoidal activation functions (Sig)  versus the same size network with (red curve) no pretraining but with batch normalization (BN) and rectified linear units (ReLU). Both networks were trained with stochastic gradient descent with momentum and mini-batches of 100. The inset clearly shows an accuracy advantage 
		of Mem-QUBO greater than $1\%$ and an error rate reduction of $20\%$ throughout the training.}
	\label{figbn}
\end{figure}

In Fig. \ref{figbn}, we see how these techniques fair against a network pretrained with the memcomputing approach on the reduced MNIST set. In the randomly initialized network, 
we employ the batch-normalization procedure \cite{pmlr-v37-ioffe15} coupled with rectified linear units (ReLUs) \cite{glorot2011deep}. As anticipated, batch-normalization smooths out the role of initial conditions, 
while rectifiers should render the energy landscape defined by Eq.~(\ref{energy}) more convex. Therefore, combined they indeed seem to provide an advantage compared to the network trained with CD using sigmoidal functions~\cite{glorot2011deep}. In fact, in experiments with batch normalization and ReLUs, we did not find any accuracy difference between a network pre-trained with CD and a randomly initialized network, and thus we omit plotting CD results in Fig.~\ref{figbn}.

However, they are not enough to overcome the advantages of our memcomputing approach. In fact, it is 
obvious from Fig. \ref{figbn} that the network pretrained with memcomputing maintains an accuracy advantage (of more than 1$\%$ and a $20\%$ reduction in error rate) on the test set out to more than a thousand back-propagation iterations. It is key to note that the network pretrained with memcomputing contains sigmoidal activations compared to the rectifiers in the network with no pretraining. Also, the pretrained network was trained without any batch normalization procedure. 

Therefore, considering all this, the pretrained network should pose a more difficult optimization problem for stochastic gradient descent. Instead, we found an accuracy advantage of memcomputing throughout the course of training. This points to the fact that with memcomputing, the pretraining procedure is able to operate close to the ``true gradient'' (Eqs.~(\ref{gradient}),~(\ref{gradientb}), and~(\ref{gradientc})) during training, and in doing so, initializes the weights and biases of the network in a advantageous way.

\section{Conclusions}
In this paper we have demonstrated how the memcomputing paradigm (and, in particular, its digital realization~\cite{dmm}) can be applied toward the chief bottlenecks in deep learning today. In this paper, we directly assisted a popular algorithm to pretrain RBMs and DBNs, which consists of gradient ascent on the log-likelihood. We have shown that memcomputing can accelerate considerably the pretraining of these networks toward better quality solutions far better than what is currently done. 

In fact, {\it simulations} of digital memcomputing machines achieve accelerations of 
pretraining comparable to, in 
number of iterations, the {\it hardware} application of the quantum annealing method, but with better quality. In addition, unlike quantum computers, our approach can be easily scaled on classical hardware to full size problems. In fact, the method we employ to solve the MAX-SAT has been shown to scale to tens of millions of variables (as opposed to only hundreds to thousands encountered in this paper)~\cite{exponential2017speedup}. We leave the study of other full size problems for future work.

In addition, our memcomputing method retains an advantage also with respect to advances in supervised training, like batch-norming and rectifiers, that have been introduced to eliminate the need of pretraining. We find indeed, that despite our pretraining done with sigmoidal functions, hence on a more non-convex landscape than that provided by rectifiers, we maintain an accuracy advantage greater than $1\%$ (and a $20\%$ reduction in error rate) throughout the training.

Finally, the form of the energy in Eq.~(\ref{energy}) is quite general and encompasses full DBNs. In this way, our method can also be applied to pretraining \emph{entire} deep-learning models at once, potentially exploring parameter spaces that are inaccessible by any other classical or quantum methods. We leave this interesting line of research for future studies.

\emph{Acknowledgments} -- We thank Forrest Sheldon for useful discussions and Yoshua Bengio for pointing out the role of supervised training techniques. H.M. acknowledges support from a DoD-SMART fellowship. M.D. acknowledges partial 
support from the Center for Memory and Recording Research at UCSD. All memcomputing simulations reported in this paper have been done using the 
Falcon$^\copyright$ simulator of MemComputing, Inc. (http://memcpu.com/).


\bibliography{haikBib}

\end{document}